\documentclass[11pt,a4paper]{article}
\usepackage[hyperref]{emnlp-ijcnlp-2019}
\usepackage{times}
\usepackage{latexsym}
\usepackage{graphicx}
\usepackage{booktabs}
\usepackage{amsmath}

\usepackage{url}

\usepackage[]{todonotes}   % insert [disable] to disable all notes
\aclfinalcopy % Uncomment this line for the final submission

\title{For Better Or Worse: On the Validation Set's Role for Early Stopping\\in Low-Resource Natural Language Processing}
\title{Towards Realistic Practices In Low-Resource Natural Language Processing: The Development Set}

\author{  
  Katharina Kann, Kyunghyun Cho and Samuel R. Bowman\\ 
  New York University, USA\\
  \texttt{\{kann, kyunghyun.cho, bowman\}@nyu.edu} 
}

\date{}

\begin{document}
\maketitle
\begin{abstract}
Development sets are 
%unlikely to be available 
impractical to obtain for real low-resource languages,
since using all available data for training is often more effective.
However, development sets are widely used %for early stopping 
in research papers that purport to deal with low-resource natural language processing (NLP). Here, we aim to answer the following questions: Does using a development set for early stopping in the low-resource setting influence results as compared to a more realistic alternative, %based on development \textit{languages}? 
where the number of training epochs is tuned on development \textit{languages}?
And does it lead to overestimation or underestimation of performance?
We repeat multiple experiments from recent work on neural models for low-resource NLP and compare results for models obtained by training with and without development sets. On average over languages, absolute accuracy differs by up to $1.4\%$.
%a more realistic alternative. 
However, for some languages and tasks, differences are as big as $18.0\%$ %absolute 
accuracy. Our results highlight the importance of realistic experimental setups 
in the publication of low-resource NLP research results. 
%for low-resource experiments in NLP.
\end{abstract}

\section{Introduction}
Parametric machine learning models are frequently trained by minimizing the loss on the training set ${\cal T}$, 
\begin{align}
  {\cal L}_{\cal T}({\boldsymbol \theta}) =& \sum_{x \in {\cal T}} l\left(\theta, x \right), \label{eq:loss}
\end{align}
for model parameters $\theta$ and a predefined loss function $l$. Gradient-based optimizers minimize this loss ${\cal L}({\boldsymbol \theta})$ by updating $\theta$ in the direction of the gradient $\nabla{\cal L}({\boldsymbol \theta})$. A low loss characterizes a model which makes accurate predictions for examples in ${\cal T}$.
%thus, ${\cal T}$ covering all datapoints the model might ever encounter would be ideal. In practice,
However, since ${\cal T}$ is finite, overfitting the training set might lead to poor generalization performance. One way to avoid fitting  Equation \ref{eq:loss} too closely is \textit{early stopping}: a separate \textit{development} or \textit{validation} set is used to end training as soon as the loss on the development set ${\cal L}_{\cal D}({\boldsymbol \theta})$ starts increasing or model performance on the development set ${\cal D}$ starts decreasing. The best 
%(not the last!) 
set of parameters ${\boldsymbol \theta}$ is used in the final model.
\begin{table}[t] % [htbp]
  \setlength{\tabcolsep}{3.2pt}
  \small
  %\normalsize
  \centering
  \begin{tabular}{ l  r r r}
    \toprule
    & \# train & \# dev & ES\\
    \midrule
    \textbf{\newcite{bollmann-etal-2018-multi}} & 5k & 12k-46k & Yes \\
    \newcite{kann-etal-2018-fortification} & 400-700 & 100-200 & Yes \\
    \newcite{makarov2018imitation} & 100 & 1k & Yes \\
    \textbf{\newcite{sharma-katrapati-sharma:2018:K18-30}} & 100 & 100 & Yes \\
    \newcite{schulz-etal-2018-multi} & 1k-21k & 9k & N/A \\
    \textbf{\newcite{upadhyay-etal-2018-bootstrapping}} & 500 & 1k & Yes \\
    \bottomrule
  \end{tabular}
  \caption{\label{tab:overview}Number of examples used for training and development in recent low-resource NLP experiments; \textit{ES}=early stopping on the development set. Experiments from papers in bold will be revisited here.
  }
\end{table}

This works well when large amounts of data are available to create training, development and test splits. Recently, however, with the success of pretraining \cite{peters-etal-2018-deep,devlin2018bert} and multi-task learning \cite{caruana1997multitask,ruder2017overview,wang2018glue} approaches, neural models   %have been gaining popularity within the natural language processing (NLP) community
are showing promising results on various natural language processing (NLP) tasks
also in \textit{low-resource} or \textit{few-shot} settings
\cite{johnson2017google,kann-etal-2017-one,yu2018diverse}. Often, the high-resource experimental setup and training procedure are kept unchanged, and 
the size of the original training set is reduced to simulate limited data.
%only the training set is being altered. 
This leads to settings where validation examples may outnumber training examples. Table \ref{tab:overview} shows such cases for the tasks of historical text normalization \cite{bollmann-etal-2018-multi}, morphological segmentation \cite{kann-etal-2018-fortification}, morphological inflection \cite{makarov2018imitation,sharma-katrapati-sharma:2018:K18-30}, argument component identification \cite{schulz-etal-2018-multi}, and transliteration \cite{upadhyay-etal-2018-bootstrapping}.    

However, in a real-world setting with limited resources, 
%as limited as implied by the training set size, 
it is unlikely that such a development set would be available for early stopping, since it would be more effective to use at least part of it for training instead. 
%all examples would be part of the training set ${\cal T}$, and, thus, not be available for early stopping. 
Here, 
we investigate how previous results relate to those obtained in a setting that does \textit{not} assume a development set.
%we investigate how experimental results change if, 
Instead of early stopping, we use data from the same task in other languages, the \textit{development languages}, to decide on the number of training epochs. We are interested in two questions: Does recent work in low-resource NLP overestimate model performance by 
using an unrealistically precise performance signal to stop training?
%stopping training at unrealistically good model parameters $\theta$? 
Or, inversely, is model performance underestimated by overfitting the finite development set?

Our experiments on %3 different 
%the tasks of 
historical text normalization, morphological inflection, and transliteration, featuring a variety of languages, show that performance does differ between runs with and without early stopping on the development set; 
%which training regime 
if using the development set 
leads to better or worse results depends on the task and language. Differences of up to $18\%$ absolute accuracy highlight that a realistic evaluation of models for low-resource NLP is crucial for estimating real-world performance.

\section{Related Work}
\paragraph{Realistic evaluation of machine learning.}
\newcite{oliver2018realistic} investigate how to evaluate semi-supervised training algorithms in a realistic way; they differ from us in that they focus exclusively on semi-supervised learning (SSL) algorithms, and do not consider NLP explicitly. However, in line with our conclusion, they report that recent practices for evaluating SSL techniques do not address the question of the algorithms' real-word applicability in a satisfying way.
In NLP, several earlier works have explicitly investigated real-world low-resource settings as opposed to artificial proxy settings, %discussed in this paper, 
e.g., for part-of-speech tagging \cite{garrette-etal-2013-real} or machine translation \cite{irvine-callison-burch-2013-combining}. While those mostly focus on real data-poor languages, we explicitly investigate the effect of the common practice to assume a relatively large development set for early stopping in the low-resource setting.

\paragraph{Low-resource settings in NLP.}
%This work can be seen as a meta-study of 
Research in the area of neural methods for low-resource NLP has gained popularity in recent years, with a dedicated workshop on the topic appearing in 2018
%titled Deep Learning Approaches for Low-Resource NLP 
\cite{ws-2018-deep}. %, whose proceedings contain many relevant papers. 
High-level key words under which other work on neural networks for data-poor scenarios in NLP can be found are domain adaptation \cite{daume-iii-2007-frustratingly}, multi-task learning \cite{caruana1997multitask,ruder2017overview}, few-shot/zero-shot/one-shot learning \cite{johnson2017google,finn2017model}, transfer learning \cite{yarowsky-etal-2001-inducing}, semi-supervised training \cite{zhu2005semi}, or pretraining \cite{erhan2010does}.

While options for early stopping without a development set exist \cite{mahsereci2017early}, they require hyperparameter tuning, which might not be feasible without a development set, and, most importantly, 
%, e.g., early stopping when the training loss converges. % cite early stopping paper! 
they are not commonly used in low-resource NLP research. Here, we investigate if \textit{current practices} might lead to unrealistic results.

\section{Experimental Design}
We compare early stopping using development set accuracy (\textit{\mbox{Dev}\mbox{Set}}) with an alternative strategy where the amount of training epochs is a hyperparameter tuned on development languages (\textit{\mbox{Dev}\mbox{Lang}}). %To this end, 
We perform two rounds of training: 
\begin{itemize}
    \item \textbf{Stopping point selection phase.} %Development language models 
    Models for the development languages are trained with the original early stopping 
    strategy from previous work.
    %which we assume available \textit{for development languages}. 
    %We record the best epoch for all languages; 
    The number of training epochs for the target languages is then calculated as the average over the best epochs for all development languages.\footnote{We round this number to an integer. It is important for our experiments that the training sets for all languages are of the same size, cf. Table \ref{tab:overview}. Otherwise we would need to account for the number of training examples per epoch.}
    %for all development languages for all non-development languages. 
    All development languages also function as target languages. To make this possible, for development languages, we compute the average over \textit{other} development languages only.
    \item \textbf{Main training phase.} We train models for all languages keeping both the model resulting from the original early stopping strategy (\mbox{Dev}\mbox{Set}) and that from the epoch computed in the stopping point selection phase (\mbox{Dev}\mbox{Lang}).\footnote{This requires training for at least the number of target epochs, even if early stopping would end training earlier.}
    %Train all models as before, but keep both the current best model and the model belonging to the epoch computed in the last step
\end{itemize}
The stopping point selection phase exclusively serves the purpose of tuning the number of epochs for the DevLang training setup. Models obtained in this phase are discarded. 
The development sets we use in our experiments are those from the original papers without alterations.

Since both final models obtained in the main training phase result from the same training run, our experimental design enables a direct comparison between the models from both setups.

\paragraph{Example.} Assume that, in the stopping point selection phase, we obtain the best development set results for a given task in development languages L1 and L2 after epochs 14 and 18, respectively. In the main training phase, we then train a model for the same task in target language L3 with the original early stopping strategy, but keeping additionally the model from epoch 16. If the best development result for language L3 is obtained after epoch 19, we compare the model from epoch 19 (DevSet) to that from epoch 16 (DevLang).

\section{Tasks, Data, Models}
For our study, we select previously published experiments which fulfill the following criteria: (1) 
datasets exist for at least four languages, and all training sets are of equal size; 
%in at least 4 languages for the same task are featured, and all training sets contain the same number of examples; 
(2) the original authors use early stopping with a development set; (3) the authors explicitly investigate low-resource settings; and (4) the original code is publically available, or a standard model is used. Since our main goal is to confirm the effect of the development set and not to compare between tasks, we further limit this study to sequence-to-sequence tasks. %All datasets and corresponding models will be described in this section. For a more detailed introduction to the neural architectures, please refer to the original papers we list.
%; models will be explained in Section \ref{sec:models}.

\subsection{Historical Text Normalization (\textsc{norm})}
\paragraph{Task.} The goal of historical text normalization is to convert old texts into a form that conforms with contemporary spelling conventions. 
%Thus, it does not aim at transforming text into \textit{another} language, but into a modern form of the \textit{same} language. 
Historical text normalization is a specific case of the general task of text normalization, which additionally encompasses, e.g., correction of spelling mistakes or normalization of social media text. %\url{https://www.aclweb.org/anthology/W18-3403}

\paragraph{Data.} We experiment on the ten datasets from \newcite{bollmann-etal-2018-multi}, which represent eight different languages: German \citep[two datasets;][]{bollmann-etal-2017-learning,odebrecht2017ridges}; English, Hungarian, Icelandic, and Swedish \cite{pettersson2016spelling}; Slovene \citep[two datasets;][]{ljubevsic2016normalising}; and Spanish and Portuguese \cite{vaamonde2015userguide}. We treat the two datasets for German and Slovene as different languages. All languages serve both as development languages for all \textit{other} languages and as target languages.

\paragraph{Model.}
Our model for this task is an LSTM \cite{hochreiter1997long} encoder-decoder model with attention \cite{bahdanau2014neural}. Both encoder and decoder have a single hidden layer. We use the default model in OpenNMT \cite{klein-etal-2017-opennmt}\footnote{\url{github.com/OpenNMT/OpenNMT-py}} 
as our implementation and employ the hyperparameters from \newcite{bollmann-etal-2018-multi}.
%Originally, 
In the original paper, \textit{early stopping} is done by training for 50 epochs, and the best model regarding development accuracy is applied to the test set.

\subsection{Morphological Inflection (\textsc{morph})}
\paragraph{Task.} Morphological inflection consists of mapping the canonical form of a word, the lemma, to an indicated inflected form. 
%An English example is \texttt{walk PAST} $\rightarrow$ \texttt{walked}. 
This task gets very complex for morphologically rich languages, where a single lemma can have hundreds or thousand of inflected forms. Recently, morphological inflection has frequently been cast as a sequence-to-sequence task, mapping the characters of the input word together with the morphological features specifying the target to the characters of the corresponding inflected form \cite{cotterell-etal-2018-conll}.

%Over the last couple of years, morphological inflection datasets in a variety of languages have been released for a series of shared tasks.
%Those shared tasks feature explicit low-resource settings, which makes the datasets natural candidates for this investigation.
\paragraph{Data.} We experiment on the datasets released for a 2018 shared task \cite{cotterell-etal-2018-conll}, which cover 103 languages and feature an explicit low-resource setting. We randomly choose ten development languages: Armenian, Basque, Galician, Georgian, Greenlandic, Icelandic, Karbadian, Kannada, Latin, and Lithuanian.

\paragraph{Model.} For \textsc{morph}, we experiment with a pointer-generator network architecture \cite{gu-EtAl:2016:P16-1,P17-1099}. This is a sequence-to-sequence model similar to that for \textsc{norm}, but employs separate encoders for characters and features. It is further equipped with a copy mechanism: using attention to decide on what element from the input sequence to copy, the model computes a probability for either copying or generation while producing an output. The final probability distribution over the target vocabulary is a combination of both. %Again, both LSTMs have 1 hidden layer. 
Hyperparameters are taken from \newcite{sharma-katrapati-sharma:2018:K18-30}.\footnote{\url{github.com/abhishek0318/conll-sigmorphon-2018}}
For \textit{early stopping}, we also follow \newcite{sharma-katrapati-sharma:2018:K18-30}: all models are trained for at least 300 epochs, and training is continued for another 100 epochs each time there has been improvement on the development set within the last 100 epochs.

\subsection{Transliteration (\textsc{transl})}
\paragraph{Task.} Transliteration is the task of converting names from one script into another, while staying as close to the original pronunciation as possible. Unlike for \textit{translation}, focus lies on the sound; the target language meaning is usually ignored.

\paragraph{Data.} For our transliteration experiments, we follow \newcite{upadhyay-etal-2018-bootstrapping}. We experiment on datasets from the Named
Entities Workshop 2015 \cite{ws-2015-named} in Hindi, Kannada, Bengali, Tamil, and Hebrew. For this task, all languages are both development and target languages.

\paragraph{Model.}
The last featured model is an LSTM sequence-to-sequence model similar to that by \newcite{bahdanau2014neural}, except for using hard monotonic attention \cite{aharoni-goldberg-2017-morphological}. 
%instead of the commonly used soft attention \cite{bahdanau2014neural}.
It attends to a \textit{single} character at a time, and attention moves monotonically over the input. %sequence. 
We take hyperparameters and code from \newcite{upadhyay-etal-2018-bootstrapping}.\footnote{\url{github.com/shyamupa/hma-translit}}
\textit{Early stopping} is done by training for 20 epochs and applying the best model regarding development accuracy to the test data.

%\textbf{Development languages.} For \textsc{norm} and \textsc{transl}, all languages are both development and target languages. For \textsc{morph}, we randomly choose 10 development languages: Armenian, Basque, Galician, Georgian, Greenlandic, Icelandic, Karbadian, Kannada, Latin, and Lithuanian.

%%% TODO: put this back in at some point %%%
\subsection{Experimental Setup} 
We run all experiments using the implementations from previous work or OpenNMT as described above. 
Existing code is only modified where necessary. Most importantly, we add storing of the DevLang model during the main training phase.

\section{Results}
\paragraph{Development sets vs. development languages. }
We are asking if the use of a development set for early stopping leads to over- or underestimation of realistic model performance. Thus, we show in Table \ref{tab:betterorworse} how often we obtain higher accuracy for each of \mbox{Dev}\mbox{Lang} and \mbox{Dev}\mbox{Set}. Additionally, averaged performance over all languages as well as the maximum difference in absolute accuracy are listed in Table \ref{tab:betterorworse_acc}. For \textsc{norm} and \textsc{transl}, results for individual languages are shown in Tables \ref{tab:det_norm} and \ref{tab:det_transl}, respectively; for detailed results for \textsc{morph} see Appendix \ref{sec:det_results}.
%Detailed results for all languages and tasks, including the respective epochs to stop training at, are in Appendix \ref{sec:det_results}.
\begin{table}[]
\small
%\normalsize
\centering
\setlength{\tabcolsep}{2.7pt}
\begin{tabular}{l| c c c}
\toprule
& \textbf{\textsc{morph}} & \textbf{\textsc{norm}} & \textbf{\textsc{transl}} \\\midrule
 \textbf{\mbox{Dev}\mbox{Lang}$>$\mbox{Dev}\mbox{Set}} & 23 & 0 & 2 \\
 \textbf{\mbox{Dev}\mbox{Lang}$=$\mbox{Dev}\mbox{Set}} & 8 & 2 & 3 \\
 \textbf{\mbox{Dev}\mbox{Lang}$<$\mbox{Dev}\mbox{Set}} & 72 & 8 & 0 \\\bottomrule
\end{tabular}
\caption{Summary of cases in which using a development set leads to overestimation (\mbox{Dev}\mbox{Set}$>$\mbox{Dev}\mbox{Lang}), underestimation (\mbox{Dev}\mbox{Set}$<$\mbox{Dev}\mbox{Lang}), or neither (\mbox{Dev}\mbox{Set}$=$\mbox{Dev}\mbox{Lang}) of the final model performance. 
%in our experiments.
\label{tab:betterorworse}}
\end{table}
We see in Table \ref{tab:betterorworse} that, for \textsc{morph} and \textsc{norm}, the use of unrealistically large development sets leads to better results than \mbox{Dev}\mbox{Lang} for 72 and 8 languages, respectively. For 8 and, respectively, 2 languages there is no difference, and a look at the detailed results in Appendix \ref{sec:det_results} and Table \ref{tab:det_norm} reveals that, for those cases, we end up training for the same number of epochs for \mbox{Dev}\mbox{Lang} and \mbox{Dev}\mbox{Set}. Only in 23 cases for \textsc{morph}, and none for \textsc{norm}, we obtain \textit{better} results for \mbox{Dev}\mbox{Lang}. This suggests that,
for these two tasks, we frequently overestimate realistic model performance by early stopping on the development set. Indeed, Table \ref{tab:betterorworse_acc} confirms this finding and shows that, on average across languages, \mbox{Dev}\mbox{Set} models outperform \mbox{Dev}\mbox{Lang} models. The maximum difference is $18\%$ absolute accuracy for the language Azeri and \textsc{morph}: for DevSet, we reach $64\%$ accuracy after epoch 217, while, for DevLang, we only obtain $46\%$ accuracy after the predefined epoch 324. 
\begin{table}[]
\small
%\normalsize
\centering
\setlength{\tabcolsep}{5.pt}
\begin{tabular}{l| c c c}
\toprule
& \textbf{\textsc{morph}} & \textbf{\textsc{norm}} & \textbf{\textsc{transl}} \\\midrule
 \textbf{\mbox{Dev}\mbox{Set}} &  \textbf{51.3} & \textbf{74.9} & 21.8 \\
 \textbf{\mbox{Dev}\mbox{Lang}} &  50.0 & 74.2 & \textbf{22.3} \\ \midrule
 \textbf{$\Delta$} & -1.4 & -0.7 & +0.5 \\ \midrule
 \textbf{max $\Delta$} & -18.0 & -2.4 & +1.3\\\bottomrule
\end{tabular}
\caption{Test accuracy in $\%$ for different stopping approaches and tasks, averaged over languages.\label{tab:betterorworse_acc}}
\end{table}

We obtain a different picture for \textsc{transl}: results are equal for 3 languages, and better for \mbox{Dev}\mbox{Lang} for the remaining 2. Equal performance might be explained by the overall smaller number of training epochs in the original regime: 
%, which reduces variance, i.e., 
stopping at the same epoch for both strategies is more likely. Overall, for \textsc{transl}, performance on the development set seems to be less predictive for the final test performance than for the other tasks. 
%This indicates that the effect of early stopping with a development set might depend on each individual validation and test set. 
\begin{table}[t] % [htbp]
  \setlength{\tabcolsep}{6.5pt}
  \small
  %\normalsize
  \centering
  \begin{tabular}{ l | c c }
\toprule
\textbf{Language} & \textbf{\mbox{Dev}\mbox{Set}} & \textbf{\mbox{Dev}\mbox{Lang}} \\\midrule
English & 0.7705 (26) & 0.7682 (43.44)\\
German (1) & 0.6749 (44) & 0.6749 (44.44)\\
German (2) & 0.7075 (43) & 0.6837 (41.78)\\
Hungarian & 0.4897 (44) & 0.4773 (41.67)\\
Icelandic & 0.7017 (27) & 0.6952 (43.56)\\
Portuguese & 0.7944 (43) & 0.7860 (42.44)\\
Slovene (1) & 0.8206 (38) & 0.8202 (41.67)\\
Slovene (2) & 0.8952 (46) & 0.8873 (42.00)\\
Spanish & 0.8352 (42) & 0.8352 (41.89)\\
Swedish & 0.7985 (43) & 0.7913 (42.11)\\
 \bottomrule
  \end{tabular}
  \caption{\label{tab:det_norm}Detailed results per language for \textsc{norm}; corresponding epochs in parenthesis.
  }
\end{table}

\begin{table}[t] % [htbp]
  \setlength{\tabcolsep}{13.pt}
  \small
  %\normalsize
  \centering
  \begin{tabular}{ l | c c }
\toprule
\textbf{Language} & \textbf{\mbox{Dev}\mbox{Set}} & \textbf{\mbox{Dev}\mbox{Lang}} \\\midrule
Bengali & 0.374 (9) & 0.387 (10.50) \\
Hebrew & 0.135 (10) & 0.135 (10.00) \\
Hindi & 0.248 (10) & 0.248 (10.25) \\
Kannada & 0.216 (11) & 0.229 (10.00) \\
Tamil & 0.116 (10) & 0.116 (10.25) \\
 \bottomrule
  \end{tabular}
  \caption{\label{tab:det_transl}Detailed results per language for \textsc{transl}; corresponding epochs in parenthesis.
  }
\end{table}

\paragraph{Influence of the final epoch. }
%TODO: Relationship between performance difference and difference in epochs
\begin{figure*}[t]
  \centering
  \includegraphics[width=1.3\columnwidth]{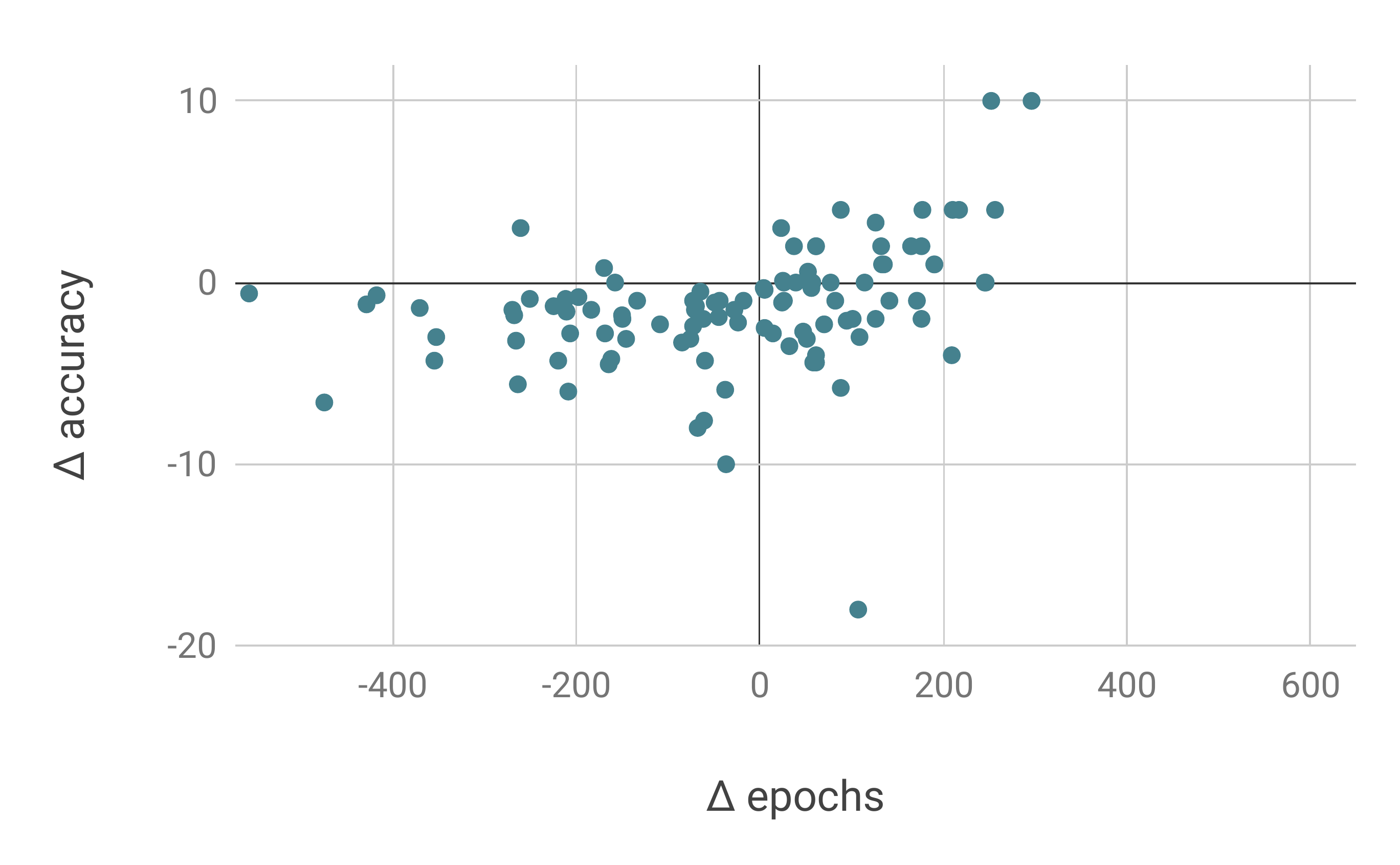}
  \caption{Difference in accuracy (\mbox{Dev}\mbox{Lang}-\mbox{Dev}\mbox{Set}) depending on the difference in training epochs (\mbox{Dev}\mbox{Lang}-\mbox{Dev}\mbox{Set}) for \textsc{morph}.}
  \label{fig:coffee}
\end{figure*}
\noindent Since without a development set performance decreases on \textsc{morph} for most languages, we investigate if this can be explained by training often being too short. Therefore, we plot the difference of training duration in epochs between \mbox{Dev}\mbox{Lang} and \mbox{Dev}\mbox{Set} against the resulting difference in accuracy in Figure \ref{fig:coffee}. While we indeed find that shorter training duration for a pointer-generator network for this task mostly results in worse performance, longer training can lead to either lower or higher accuracy. Thus, while longer training seems better for \textsc{morph}, not all performance loss can be explained by aborting training too early.
%, but that, if in doubt, longer training is recommendable for \textsc{morph}.

%\subsection{Development Vs. Test Performance}

%However, most of these work do not fulfill the requirements listed in the introduction and, thus, are not featured in this study.

\section{Discussion And Conclusion}
\paragraph{Limitations.}
We investigate the effect of early stopping on the validation set
%on experimental results 
as compared to a realistic setting without target language development examples.
%Thus, our goal is to explore differences between frequent assumptions and potential real-world settings. 
%However, 
However, we would like to point out that, in certain situations, standard practices might be sufficient, e.g., for comparing different methods in equal settings, if absolute performance is not the main focus.

Further, %we emphasize that 
we do not claim to show that using a validation set {always} \textit{over}- or \textit{under}estimates real-world performance, since this %most likely 
depends on how representative the validation set is of the target distribution. Our main result is that using a development set gives a poor estimate of real-world performance and that it is important to be aware of potential performance differences. 

 %Additionally,
%they require the introduction of new hyperparameters ("patience"), which, in turn, requires a validation set or development languages. 
%Thus, we exclude them here for a clean comparison, but 
%and might be worth studying %as an alternative to \mbox{Dev}\mbox{Lang} in future work. 

\paragraph{Practical take-aways.}
%In this work, 
We replicate experiments from recent low-resource NLP research, once with the original experimental design and once without assuming development sets for early stopping. 
Since differences in absolute accuracy are up to $18.0\%$, we
%Using the development set leads to significant differences in model performance as compared to an alternative without it. Thus, 
conclude that %the most important conclusion from this study is that 
low-resource NLP research should move away from using large development sets for early stopping whenever real-world settings are being considered. 
%Development languages might be a suitable alternative. 
%One possible alternative are development languages.
%researchers should stop relying on it for early stopping if low-resource settings are the subject of investigation. 

\section*{Acknowledgments}
We would like to thank Nikita Nangia and Jason Phang for their feedback on this work.
This work has benefited from the support of Samsung Research under the project \textit{Improving Deep Learning using Latent Structure}
and from the donation of a Titan V GPU by NVIDIA Corporation.

\bibliography{emnlp-ijcnlp-2019}
\bibliographystyle{acl_natbib}
\vfill
\pagebreak
~
\vfill
\pagebreak
\appendix
\section{Detailed Results for \textsc{morph}}
\label{sec:det_results}
\vspace{3cm}
\begin{table}[h] % [htbp]
  \setlength{\tabcolsep}{6.7pt}
  \scriptsize
  \centering
  \begin{tabular}{ l | c c }
\toprule
\textbf{Language} & \textbf{\mbox{Dev}\mbox{Set}} & \textbf{\mbox{Dev}\mbox{Lang}} \\\midrule
adyghe & 89.2 (263) & 84.8 (324)\\
albanian & 27.9 (400) & 24.8 (324)\\
arabic & 34.1 (397) & 31.7 (324)\\
armenian & 52.2 (393) & 50.9 (323.1111111)\\
asturian & 71.2 (348) & 69 (324)\\
azeri & 64 (217) & 46 (324)\\
bashkir & 80.1 (474) & 78.1 (324)\\
basque & 7.9 (282) & 8.5 (334.2222222)\\
belarusian & 22.1 (198) & 25.4 (324)\\
bengali & 56 (585) & 59 (324)\\
breton & 56 (263) & 58 (324)\\
bulgarian & 50.3 (266) & 45.9 (324)\\
catalan & 64 (319) & 61.5 (324)\\
classical-syriac & 93 (285) & 93 (324)\\
cornish & 32 (267) & 32 (324)\\
crimean-tatar & 90 (458) & 89 (324)\\
czech & 37.2 (592) & 35.4 (324)\\
danish & 64.5 (409) & 61.2 (324)\\
dutch & 54.1 (362) & 48.2 (324)\\
english & 87.3 (292) & 83.8 (324)\\
estonian & 31 (536) & 30.1 (324)\\
faroese & 34 (310) & 31.2 (324)\\
finnish & 21.5 (799) & 14.9 (324)\\
french & 55.1 (881) & 54.5 (324)\\
friulian & 72 (247) & 72 (324)\\
galician & 44 (230) & 41.9 (324.4444444)\\
georgian & 77.2 (353) & 76.1 (303.5555556)\\
german & 52.3 (384) & 48 (324)\\
greek & 25.6 (300) & 24.5 (324)\\
greenlandic & 74 (167) & 76 (331.6666667)\\
haida & 58 (298) & 57 (324)\\
hebrew & 31.3 (268) & 31 (324)\\
hindi & 75 (594) & 73.5 (324)\\
hungarian & 39.5 (395) & 38 (324)\\
icelandic & 33.9 (250) & 31.6 (319.8888889)\\
ingrian & 50 (263) & 46 (324)\\
irish & 23.7 (493) & 20.9 (324)\\
italian & 45.6 (549) & 44.3 (324)\\
kabardian & 89 (233) & 86 (341.4444444)\\
kannada & 52 (93) & 56 (349.2222222)\\
karelian & 88 (147) & 92 (324)\\
kashubian & 52 (263) & 54 (324)\\
kazakh & 82 (287) & 84 (324)\\
khakas & 82 (148) & 80 (324)\\
khaling & 23.1 (742) & 22.4 (324)\\
kurmanji & 81.9 (470) & 78.8 (324)\\
ladin & 66 (236) & 70 (324)\\
latin & 14.4 (478) & 10.2 (315.8888889)\\
latvian & 35.3 (397) & 34.3 (324)\\
lithuanian & 15.8 (447) & 14 (296.5555556)\\
livonian & 31 (134) & 32 (324)\\
 \bottomrule
  \end{tabular}
  \caption{\label{tab:det_morph}Detailed results per language for \textsc{morph}; corresponding epochs in parenthesis; part 1.
  }
\end{table}

\begin{table}[t] % [htbp]
  \setlength{\tabcolsep}{6.7pt}
  \scriptsize
  \centering
  \begin{tabular}{ l | c c }
\toprule
\textbf{Language} & \textbf{\mbox{Dev}\mbox{Set}} & \textbf{\mbox{Dev}\mbox{Lang}} \\\midrule
lower-sorbian & 42 (590) & 38.8 (324)\\
macedonian & 56.8 (508) & 55.3 (324)\\
maltese & 26 (189) & 27 (324)\\
mapudungun & 86 (107) & 90 (324)\\
middle-french & 82.9 (236) & 77.1 (324)\\
middle-high-german & 80 (223) & 78 (324)\\
middle-low-german & 36 (192) & 38 (324)\\
murrinhpatha & 38 (72) & 48 (324)\\
navajo & 10.8 (368) & 9.8 (324)\\
neapolitan & 77 (301) & 80 (324)\\
norman & 60 (148) & 62 (324)\\
northern-sami & 18.7 (695) & 17.3 (324)\\
north-frisian & 43 (386) & 41 (324)\\
norwegian-bokmaal & 66.3 (273) & 63.2 (324)\\
norwegian-nynorsk & 51.6 (385) & 44 (324)\\
occitan & 69 (183) & 68 (324)\\
old-armenian & 31.6 (389) & 31.1 (324)\\
old-church-slavonic & 49 (134) & 50 (324)\\
old-english & 22.7 (433) & 20.4 (324)\\
old-french & 40.6 (352) & 39.1 (324)\\
old-irish & 2 (78) & 2 (324)\\
old-saxon & 25 (494) & 25.8 (324)\\
pashto & 36 (114) & 40 (324)\\
persian & 51.2 (277) & 48.5 (324)\\
polish & 31.3 (544) & 27 (324)\\
portuguese & 57.3 (319) & 56.9 (324)\\
quechua & 58.7 (753) & 57.5 (324)\\
romanian & 35.2 (531) & 32.4 (324)\\
russian & 43.6 (320) & 43.3 (324)\\
sanskrit & 50.7 (489) & 46.2 (324)\\
scottish-gaelic & 76 (361) & 66 (324)\\
serbo-croatian & 34.4 (533) & 28.4 (324)\\
slovak & 41 (575) & 40.1 (324)\\
slovene & 47.4 (535) & 45.8 (324)\\
sorani & 27.3 (677) & 24.3 (324)\\
spanish & 54.3 (588) & 48.7 (324)\\
swahili & 62 (298) & 62 (324)\\
swedish & 63.7 (392) & 55.7 (324)\\
tatar & 76 (191) & 77 (324)\\
telugu & 98 (79) & 98 (324)\\
tibetan & 36 (28) & 46 (324)\\
turkish & 36.6 (299) & 36.7 (324)\\
turkmen & 82 (115) & 78 (324)\\
ukrainian & 33.8 (522) & 33 (324)\\
urdu & 66.6 (369) & 64.7 (324)\\
uzbek & 93 (242) & 92 (324)\\
venetian & 76.8 (210) & 76.8 (324)\\
votic & 22 (342) & 21 (324)\\
welsh & 48 (482) & 48 (324)\\
west-frisian & 50 (153) & 49 (324)\\
yiddish & 63 (198) & 61 (324)\\
zulu & 32.3 (679) & 28 (324)\\
 \bottomrule
  \end{tabular}
  \caption{\label{tab:det_morph2}Detailed results per language for \textsc{morph}; corresponding epochs in parenthesis; part 2.
  }
\end{table}
\end{document}